\documentclass{article}
\usepackage{ijcai25}
\usepackage{marvosym}
\usepackage{xspace}
\usepackage{times}
\usepackage{soul}
\usepackage{url}
\usepackage[hidelinks]{hyperref}
\usepackage[utf8]{inputenc}
\usepackage[small]{caption}
\usepackage{graphicx}
\usepackage{amsmath}
\usepackage{amsthm}
\usepackage{booktabs}
\usepackage{algorithm}
\usepackage{algorithmic}
\usepackage[switch]{lineno}
\usepackage{tcolorbox}

\usepackage{booktabs}        
\usepackage{array}           
\usepackage{multirow}        
\usepackage{multicol}        
\usepackage{adjustbox}       

\usepackage{color}           
\usepackage{xcolor}          
\usepackage{textcomp}        

\usepackage{amsmath}         
\usepackage{amssymb}         
\usepackage{bm}              

\usepackage{caption}         
\usepackage{subcaption}      
\usepackage{graphicx}        

\newcommand{\datasetFont}{\text}
\newcommand{\ours}{\datasetFont{DeepShade}\xspace}

\title{DeepShade: Enable Shade Simulation by Text-conditioned Image Generation}

\author{
Longchao Da$^*$
\and
Xiangrui Liu$^*$\and
Mithun Shivakoti\and
Thirulogasankar Pranav Kutralingam\and
Yezhou Yang\And
Hua Wei$^{\dag}$\\
\affiliations
Arizona State University
\emails
\{longchao, xiangrui, mshivako, tknolast, yz.yang, hua.wei\}@asu.edu
}

\begin{document}

\maketitle

\begin{abstract}
    Heatwaves pose a significant threat to public health, especially as global warming intensifies. However, current routing systems (e.g., online maps) fail to incorporate shade information due to the difficulty of estimating shades directly from noisy satellite imagery and the limited availability of training data for generative models. In this paper, we address these challenges through two main contributions. First, we build an extensive dataset covering diverse longitude-latitude regions, varying levels of building density, and different urban layouts. Leveraging Blender-based 3D simulations alongside building outlines, we capture building shadows under various solar zenith angles throughout the year and at different times of day. These simulated shadows are aligned with satellite images in terms of the areas, providing a rich resource for learning shade patterns. Second, we propose the DeepShade, a diffusion-based model designed to learn and synthesize shade variations over time. It emphasizes the nuance of edge features by jointly considering RGB with the Canny edge layer, and incorporates contrastive learning to capture the temporal change rules of shade. Then, by conditioning on textual descriptions of known conditions (e.g., time of day, solar angles), our framework provides improved performance in generating shade images. We demonstrate the utility of our approach by using our shade predictions to calculate shade ratios for real-world route planning in Tempe, Arizona. We hope this work could provide a reference for urban planning in extreme heat weather and reveal its potential practical applications in the environment.
\end{abstract}

\section{Introduction}
\label{sec:intro}


Extreme weather is causing an increasing number of deaths worldwide, with heatwaves being a major contributing factor. According to a report~\cite{casereport}, the frequency and intensity of extreme heat events have surged over the past two decades, with more than 178,700 deaths occurring annually (average from 2000 to 2019) as a direct result of high temperatures~\cite{monashstudy}. Research from the World Health Organization highlights that extreme heat is now one of the leading causes of weather-related deaths~\cite{niehsheatimpact}, disproportionately affecting vulnerable populations such as the elderly and those working or staying in outdoor areas with limited access to cooling infrastructure. This trend underscores the urgent need for adaptive measures, including heat-resilient urban design and shade-aware route-planning methods~\cite{ma2018parasol}, to mitigate the public health impact of rising temperatures globally. 

Since shade acts as a natural shelter that reduces direct exposure to solar radiation, understanding how shade changes in real-time is crucial for preparing outdoor activities and aiding urban planning in establishing artificial shelters in areas lacking natural shade. As introduced in~\cite{da2024shaded}, by identifying shaded areas and integrating this information into route-planning systems, individuals can make more comfortable travel choices, reducing their risk of heat-related illnesses. However, the existing research faces significant limitations: \textbf{First}, shade analysis using urban simulations is only on static maps, which lacks good generalizability. \textbf{Second}, most of the methods are localized, relying on resource-intensive LiDAR data; they also lack scalability across different or more extensive areas.  \textbf{Third}, they are unable to capture the real-time shade dynamics, limiting their utility in intelligent routing.  These limitations impede the development of accurate shade modeling and timely planning, affecting the practical impact of existing studies.

To tackle the above challenges, a more adaptive and scalable approach to dynamically model shade variations is required. Given the success of Generative AI~\cite{da2024prompt} (particularly diffusion models), in capturing spatial-temporal patterns in urban scene synthesis~\cite{tang2024diffuscene} and environment simulation~\cite{da2025survey}, they present a promising direction for overcoming these challenges by generating shades for satellite images. However, to effectively leverage this method, a well-structured dataset, that accurately aligns ground-truth shade-variance with its satellite-image level geographic information, is necessary. Such a dataset is left blank in the current research domain. 

\begin{figure*}[t!]
    \centering
    \includegraphics[width=1\textwidth]{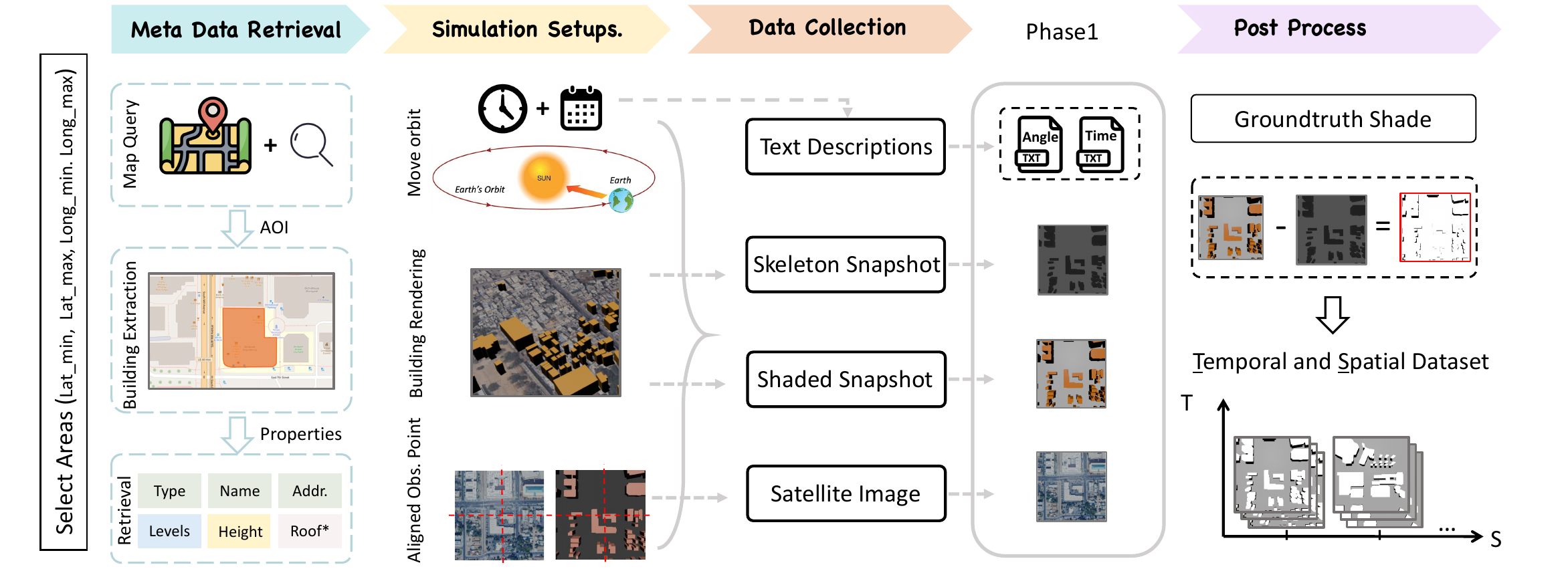}
    \caption{The overview of the pipeline, note that `Addr.' is an abbreviation of Address, and `Roof*' includes: roof shape and roof height, the building shape is approximated by the bird’s-eye view of the roof shape. 
    }
    \label{fig:pipeline}
\end{figure*}

In this paper, we \textbf{first} developed a rigorous and systematic pipeline to construct a comprehensive dataset for training shade-generation models. The dataset is designed to encompass three critical dimensions while adhering to a unified standard. The three dimensions include: (1) Geographical Diversity, covering a wide range of continents with varying latitudinal and longitudinal distributions; (2) Urban Layout Variability, capturing diverse urban configurations such as dense high-rise building places and sparse flat areas; and (3) Traffic Rule Variation, accounting for differences in driving orientations, including left-hand and right-hand traffic systems. The unified standard is that the cities must suffer from \textbf{significant heat events} as defined by the World Meteorological Organization (WMO), which guarantees the dataset focuses on regions with high temperatures and their associated impacts on shade dynamics are most pronounced. We \textbf{secondly}, designed a novel contrastive learning-based diffusion model approach, with a fine-granularity edge conditional module, that learns the shade variations based on the skeleton representations of corresponding satellite images. The model effectively maps shade dynamics to text prompts containing temporal and geographic information, allowing it to generate realistic shade predictions for arbitrary times of day or specific solar angles. This solution is unique for its ability to train on a limited satellite image dataset while generalizing to unseen buildings, provided a satellite image is available. This relaxed the requirements for detailed building height to conduct simulation and ensures adaptability and robustness across diverse urban settings.

In conclusion, this work contributes to advancing shade-generation research by addressing critical challenges in data preparation, model design, and real-world application. \textbf{First,} we developed a comprehensive, globally representative dataset, meticulously crafted to align satellite imagery with the dataset. \textbf{Second,} we introduced a text-conditioned image generation model leveraging edge conditioning and contrastive learning, enabling accurate and generalizable shade predictions. \textbf{Third,} we conducted extensive experiments across diverse cities, showcasing the model's robustness in handling varying landscapes, urban layouts, and geospatial features. \textbf{Finally,} we demonstrated a practical application of the model in shaded route planning, where it generates shade maps for different times of the day based on satellite images and textual prompts. These contributions collectively provide a robust framework for urban planning and heat mitigation strategies.

\section{Related Work}
\subsection{Building Shade Analysis}

There are works investigating the shade changes from buildings and how they affect people’s living. A recent study~\cite{park2023quantifying}, has quantified the cumulative cooling effects in urban environments, particularly emphasizing the benefit of high-resolution thermal imagery in hot, arid climates like Tempe, Arizona.  Another recent work also used high-resolution data and models like SOLWEIG~\cite{lindberg2008solweig} to map the thermal impact of shade at fine scales, offering valuable insights for urban planners. These maps are used in initiatives such as ``cool corridor" planning~\cite{buo2023high} and ~\cite{wu2020construction}, which aims to reduce heat exposure for pedestrians by optimizing sidewalk shade coverage. Such analysis highlights the focus on reducing urban heat in the modern environment~\cite{buo2023high}. However, current research typically focuses on specific urban locations and relies heavily on high-resolution LiDAR data, which, although accurate, is costly and complex to collect on a city-wide large scale. This has led to our work in developing universally applicable shade analysis methods that could comprehensively assess shade impacts across diverse cities.

\subsection{City Simulation}

There exist studies~\cite{kamath2024global} that utilized high-resolution datasets, like the University of Texas Global Building Heights for Urban Studies (UT-GLOBUS~\cite{li2020developing}). UT-GLOBUS combines data from sources such as LiDAR, ICESat-2~\cite{abdalati2010icesat}, and Global Ecosystem Dynamics Investigation (GEDI)~\cite{schneider2020towards}, augmented by machine learning models, to produce comprehensive canopy information for urban studies. 
Existing simulations like those conducted using GIS in the Adab neighborhood of Sanandaj~\cite{beheshtifar2023simulation} demonstrate the complexity of modeling urban shade in dense areas with high-rise structures. The study used 3D GIS data to simulate shade across different seasons, highlighting the intense shading experienced by 75\% of buildings in winter. However, all of the above datasets and simulations remain limited to specific locations and lack the scalability to encompass up-to-date, diverse urban landscapes or support universal shade simulations across varying city layouts and climates.

\subsection{Generative Models}
Generative models~\cite{da2025generative} such as diffusion models~\cite{croitoru2023diffusion} (used in image synthesis~\cite{dhariwal2021diffusion} and refinement~\cite{du2023arsdm}), ControlNet~\cite{zhao2024uni}, and StyleGAN~\cite{abdal2019image2stylegan} excel at producing detailed synthetic data. ControlNet uses external signals for spatial guidance, and StyleGAN enables high-resolution, realistic image generation with fine-grained style control~\cite{azadi2018multi}. However, these models have yet to be applied to urban analysis (e.g., shade generation) due to challenges in acquiring accurately calibrated, geo-referenced shade data that matches satellite imagery and models sunlight angles precisely. In this work, we propose a novel framework to address this by constructing a comprehensive, calibrated shade dataset with ground truth shades.

\section{Dataset Creation}
The archived geographical information might be out-of-date which hinders the feasibility of simulating shade changes, but it is possible to use accessible `satellite image + generalizable models' to infer the shade areas.

In order to develop generalizable models, we construct a formal pipeline that creates a comprehensive dataset covering three dimensions: Geographical Diversity, Urban Layout Variability, and  Traffic Rule Variation. This dataset aligns satellite images with Open Street Map (OSM)~\cite{map2017open} building information. The pipeline involves four major steps as shown in Figure~\ref{fig:pipeline}, which illustrates the overall framework from input to the collected outcome.

First, \texttt{Metadata Retrieval} obtains all necessary metadata used for the 3-D simulation. To align with the research community and utilize the large open-source data, we adopt OSM data as the metadata for shade simulation. Moreover, based on the longitude and latitude information, we can extract the building's geographical information, including ``property type, address, height, levels, shape, and geometry set of locations'', this typically provides information such as a list of points that form a polygon of building's 3D skeleton.

Second, \texttt{Simulation Setups}. With the necessary data collected, we performed the simulation to capture the shade changes for our dataset creation. In addition, we use Blender~\cite{hess2013blender} for large-scale city-wide shade simulation. Based on scientific rules, we set up a controller as the sun's movement engine, which can generate the sun's trajectory following the sun's solar declination or angle on any day of the year, and at any time of the day. Then, we rendered the area of interest (AOI) within the simulation platform and adjusted the scale to ensure it aligns with the prevalent maps. Here, we align images with Google map tile level \textcolor{black}{13}. 

Third, \texttt{Data Collection}. We collected four types of data: (1) skeleton snapshots, (2) shaded snapshots, (3) satellite images, and (4) text descriptions. These diverse data representations enable learning models to understand shade variations and their dependencies on environmental factors. 

The \textbf{shaded snapshot} \( x^{shade} \) is obtained with parallel sunshine casting on the scene, capturing shades, side projections, and full structural visibility of buildings. To extract a structure-only reference, we generate a \textbf{skeleton snapshot} \( x^{sk} \) by rendering the identical scene with the sun being invisible. Additionally, a \textbf{satellite image} \( x^{sat} \) of the corresponding area is aligned the simulated environment with real-world conditions. Finally, we extract \textbf{text descriptions} \( T \) encoding temporal information, such as the sun angle and time of day, which provide an additional modality for shade generation control. We formulate the text description as follows:
\begin{equation}
    T = f(\theta_{sun}, t_{day})
\end{equation}
where \( T \) represents the generated text prompt, \( \theta_{sun} \) denotes the solar declination or angle, and \( t_{day} \) is the timestamp. The function \( f(\cdot) \) transforms solar and temporal attributes into natural language descriptions, which is a simple string combination function with alternative descriptive choices. 
\noindent The following example showcases structured text prompts for different temporal conditions:

\begin{tcolorbox}[colback=gray!10, 
                  colframe=black, 
                  width=8.5cm, 
                  arc=1mm, auto outer arc,
                  boxrule=0.05pt,
                  fontupper=\small] 

\textbf{Example Text Prompts:} \\[2pt]
\textbf{Prompt 1:} Solar declination: -20.7° \\ 
\textbf{Prompt 2:} Angle: 45° \\ 
\textbf{Prompt 3:} Right now, it is 6:00 PM in a day. 

\end{tcolorbox}


Last, we have the \texttt{Post process} step. To benefit the deep learning models, we need a ground-truth dataset for measuring the model's output and providing feedback through loss functions. However, current data from step 3 is not the ultimate shade as it includes side shadows and noises. Therefore, to obtain the actual ground truth shade, we propose to take another snapshot of the scene without sunshine, named as `skeleton image' $x^{sk}$, which only provides a pure building structure, then the pure shade can be obtained through the formula:
\begin{equation}\label{eq:threhsold}
    x^{gt} = x^{shade} - x^{sk} - \mathbb{I}(x^{shade} \leq \alpha)
\end{equation}
where \( x^{gt} \) represents the extracted ground truth shade, \( x^{shade} \) is the shaded snapshot taken under sunlight, capturing the shades, sides, and full building observation, and \( x^{sk} \) is the skeleton snapshot taken without sunlight, providing only the pure building structure. The term \( \mathbb{I}(x^{shade} \leq \alpha) \) is an indicator function that removes low-intensity grey values (noise thresholding), where \( \alpha \) is a predefined threshold. 
This equation effectively removes structural elements and noise, isolating the actual shaded areas to provide precise ground truth data for machine learning models.





The dataset construction process has improved alignment between simulated and real-world data while enabling controllable shade generation via images, textual descriptions, or a joint approach.
This dataset has been split into train, test sets by 70\%, 30\%, which can help the research community with a well-defined benchmark (researcher could hold another validation set by changing the index if necessary).

\section{A Text Conditioned Shaded Image Generation Model}
\begin{figure*}[h!]
    \centering
    \includegraphics[width=0.98\linewidth]{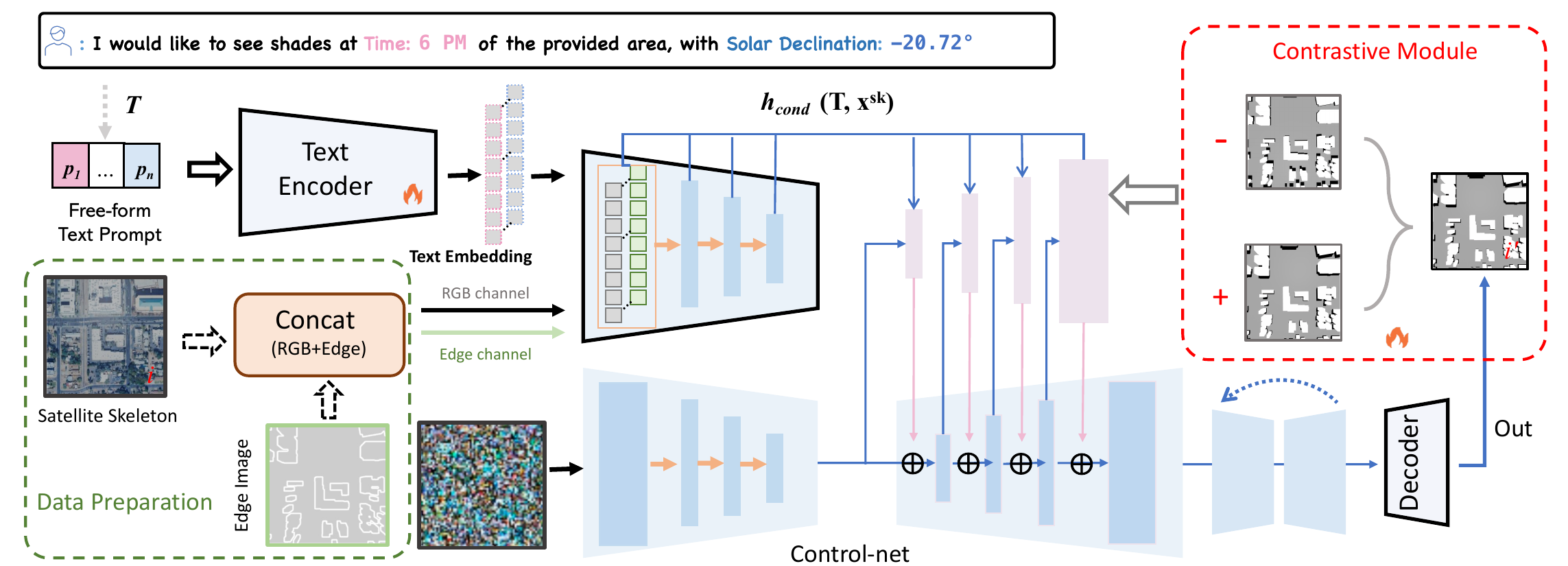}
    \caption{The structure of the proposed DeepShade method, based on the basic control net, we design two parts, \textbf{first} one on the left hand is the data preparation module, which concatenates the skeleton image with the Canny edge features, it helps the model focus on the shade edges and capture the overall skeleton of the building structures. The \textbf{second} on the right part is the contrastive part, the constructed contrastive buffer takes positive and negative pairs, during the training period, it optimizes the model by comparing the generated image $i'$ with two contrastive pairs, and effectively learns the temporal difference reflected on the edge features.}
    \label{fig:mainfig1}
\end{figure*}

Now given a well-constructed shade dataset $D = \{ x^{shade}, x^{sk}, x^{sat}, T \}$, where \( x^{shade} \) is the shaded snapshot capturing shadowed areas under sunlight, \( x^{sk} \) is the skeleton snapshot, \( x^{sat} \) is the real-world satellite image after alignment, and \( T \) is the textual description encoding solar angle \( \theta_{sun} \) and timestamp \( t_{day} \). To benefit from that, this paper also explores methods that generate accurate shade images following the text prompt and base image (which provides a base map for generation).  Given the objective is to generate shade images using a base map and text conditions, we adopt a diffusion-based framework, ControlNet~\cite{zhang2023adding}, as our backbone since the diffusion models are effective in image synthesis and editing tasks~\cite{dhariwal2021diffusion}.
As shown in Figure~\ref{fig:mainfig1}, our proposed method focuses on \textit{edge information incorporation} and \textit{contrastive learning module}. We will introduce the details after a reflection on the Controlnet Model in our setting.

\paragraph{Reflection on ControlNet in Our Task}
ControlNet is an extension of diffusion models designed to incorporate visual conditions into the image generation process. It is built upon pre-trained Stable Diffusion~\cite{blattmann2023stable} models by introducing an additional control pathway that enables precise guidance during generation. 

In our task, the vanilla ControlNet can be used to generate shade images conditioned on both structural and descriptive inputs (text description of time/angle). The base map \( x^{sk} \), representing the skeleton structure of the scene, serves as the primary visual condition. The text prompt \( T \), which encodes solar angle \( \theta_{sun} \) and timestamp \( t_{day} \), provides semantic and temporal guidance. These inputs collectively form the conditioning inputs \( c \), driving the generation process. The generation process can be expressed as:
\begin{equation}
    x^I = G(T, x^{sk})
\end{equation}

where \( x^I \) is the generated shade image, \( T \) is the text prompt, \( x^{sk} \) is the base map. At each layer \( i \) of the decoder \( D \), ControlNet integrates these conditions as:
\[
\mathbf{h}^{i+1} = D^i(\mathbf{h}^i + \mathbf{h}^i_{\text{cond}})
\]

where \( \mathbf{h}^i \) represents the feature map at layer \( i \), and \( \mathbf{h}^i_{\text{cond}} \) is the feature derived from the combined conditions \( c \). This mechanism ensures that the generated output adheres to both the structural layout (base map) and temporal dynamics (descriptions) provided by the inputs.

The vanilla ControlNet framework could functionally generate shade images, but the quality of generation is limited, mainly because of two reasons: 

\textbf{\textit{First: The shade generation involves very subtle features}}: Shade occupies only a minor part of an image and is often nuanced and difficult to capture.  Thus, identifying and processing these subtle features requires a further refinement of the model's design.

\textbf{\textit{Second, The shade changes follow temporal rule}}: Shade changes dynamically over time, following natural rules related to solar angles and time progression. Also, the shade changes can be subtle in close timing. The vanilla ControlNet struggles to accurately simulate these temporal dependencies, often generating the wrong orientation of shades due to the failure to capture the consistent and realistic evolution of shade in response to these factors.

\subsection{Edge‐Enhanced Conditional Generation}  
To address the first challenge that the vanilla model fails to capture subtle shade outline features and often generates shadows with crooked boundaries, even for buildings with straight outlines, we propose an edge‐enhanced conditional generation method. The approach incorporates edge information as an additional visual condition alongside the base map, as shown in the preparation part of the Figure~\ref{fig:mainfig1}.  

Specifically, our base map \(x^{sk}\in\mathbb{R}^{H\times W\times 3}\) is a three‐channel mask of building footprints. We extract structural boundaries via Canny edge detection to obtain a single‐channel edge map \(x^{edge}\in\mathbb{R}^{H\times W\times 1}\). These are concatenated along the channel dimension to form a four‐channel conditioning tensor (R, G, B and edge): 
\begin{equation}
    x^{cond}
=\bigl[x^{sk}_{R},\,x^{sk}_{G},\,x^{sk}_{B},\,x^{edge}_{}\bigr]
\;\in\;\mathbb{R}^{H\times W\times 4}.
\end{equation}

This four‐channel tensor is supplied to the ControlNet's U-Net as its concatenated input, together with the text prompt \(T\). By providing both the global building layout and its precise edge contours, the generator is shown below:  
\begin{equation}\label{eq:input}
x^I \;=\; G\bigl(T,\;x^{cond}\bigr)    
\end{equation}
It is guided to produce shadows with straight, well‐aligned boundaries.

\subsection{Contrastive Based Shade Generation}

To tackle the second challenge, we innovatively propose a contrastive learning enforced learning paradigm for shade generation. We first construct a contrastive data pipeline, introduce a lightweight contrastive encoder to embed the features, and then introduce the contrastive learning-based shade generation.

\newcolumntype{P}[1]{>{\centering\arraybackslash}p{#1}}
\begin{table*}[thb!]
\scriptsize
    \centering
    \begin{tabular}{P{2.7cm}|P{0.7cm}P{0.7cm}|P{0.7cm}P{0.7cm}|P{0.7cm}P{0.7cm}|P{0.7cm}P{0.7cm}|P{0.7cm}P{0.7cm}|P{0.7cm}P{0.7cm}}\toprule
        \multirow{2}{*}{\textbf{Baselines on Scenario1 }}   & \multicolumn{2}{c|}{Beijing (CHN)} & \multicolumn{2}{c|}{Phoenix (USA)} & \multicolumn{2}{c|}{São Paulo (BRA)} & \multicolumn{2}{c|}{Madrid (ESP)}& \multicolumn{2}{c|}{Cairo (EGY)} & \multicolumn{2}{c}{Mumbai (IND)}\\ 
        \cmidrule(lr){2-3} \cmidrule(lr){4-5} \cmidrule(lr){6-7} \cmidrule(lr){8-9} \cmidrule(lr){10-11} \cmidrule(lr){12-13} 
         &  SSIM\(\uparrow\) &  LPIPS\(\downarrow\) & SSIM\(\uparrow\) &  LPIPS\(\downarrow\) &   SSIM\(\uparrow\) & LPIPS\(\downarrow\) &  SSIM\(\uparrow\) &  LPIPS\(\downarrow\) &  SSIM\(\uparrow\) &  LPIPS\(\downarrow\) & SSIM\(\uparrow\) &  LPIPS\(\downarrow\) \\ \midrule
         Diffusion Model & 0.610 & 0.518 & 0.411 & 0.446 & 0.475 & 0.440  & 0.388&0.417&0.357& 0.437 & 0.352 & 0.399\\
         ControlNet &  0.941 & 0.225 & 0.941 &  0.265 & 0.951 & 0.291 & 0.936 & 0.277 & 0.944 & 0.265 & 0.915 & 0.254\\
         Edge Control &  0.934 & 0.225 & 0.934 &  0.254 &0.954 & 0.284 & 0.946  & 0.243  & 0.942  & 0.267  & 0.929  & 0.273 \\
         \textbf{\ours{}} & \textbf{0.945} & \textbf{0.194} & \textbf{0.946} & \textbf{0.164} & \textbf{0.959} & \textbf{0.210}  & \textbf{0.948} & \textbf{0.239} & \textbf{0.954} & \textbf{0.257} & \textbf{0.931} & \textbf{0.250}\\ 
        \bottomrule
    \end{tabular}
    \begin{tabular}{P{2.7cm}|P{0.7cm}P{0.7cm}|P{0.7cm}P{0.7cm}|P{0.7cm}P{0.7cm}|P{0.7cm}P{0.7cm}|P{0.7cm}P{0.7cm}|P{0.7cm}P{0.7cm}}\toprule
        \multirow{2}{*}{\textbf{Baselines on Scenario2}}   & \multicolumn{2}{c|}{Xi'An (CHN)} & \multicolumn{2}{c|}{Tempe (USA)} & \multicolumn{2}{c|}{Brasilia (BRA)} & \multicolumn{2}{c|}{Seville (ESP)}& \multicolumn{2}{c|}{Aswan (EGY)} & \multicolumn{2}{c}{Jaipur (IND)}\\ 
        \cmidrule(lr){2-3} \cmidrule(lr){4-5} \cmidrule(lr){6-7} \cmidrule(lr){8-9} \cmidrule(lr){10-11} \cmidrule(lr){12-13} 
         &  SSIM\(\uparrow\) &  LPIPS\(\downarrow\) & SSIM\(\uparrow\) &  LPIPS\(\downarrow\) &   SSIM\(\uparrow\) & LPIPS\(\downarrow\) &  SSIM\(\uparrow\) &  LPIPS\(\downarrow\) &  SSIM\(\uparrow\) &  LPIPS\(\downarrow\) & SSIM\(\uparrow\) &  LPIPS\(\downarrow\) \\ \midrule
         Diffusion Model & 0.425 & 0.468 & 0.381& 0.347& 0.402& 0.400 & 0.342 & 0.434 & 0.356& 0.456 & 0.339& 0.357\\
         ControlNet & 0.930 & 0.236 & 0.969 & 0.330 & 0.973 & 0.297 &0.934&0.281&0.967& 0.352 & 0.948 & 0.297\\
         Edge Control &  0.930  & 0.237  & 0.968 &  0.335 & 0.967 & 0.299 & 0.936 & 0.276 & 0.966 & 0.379 & 0.947 & \textbf{0.273}\\
         \textbf{\ours{}} & \textbf{0.932} & \textbf{0.233} & \textbf{0.969} & \textbf{0.329} & \textbf{0.979} & \textbf{0.241} & \textbf{0.936} & \textbf{0.242} & \textbf{0.973} & \textbf{0.283} & \textbf{0.956} & 0.275 \\ 
        \bottomrule
    \end{tabular}
    \caption{\textbf{Test results from two scenario types (dense\textbf{*} and sparse\textbf{*} building cities)}: \ours{} consistently outperforms other baselines on both experimental scenarios, demonstrating superior language understanding and segmentation accuracy across in-domain and out-of-domain datasets, even when applied with random transformations. $\uparrow$ means the larger, the better, while $\downarrow$ means the lower is more expected.}
    \label{tab:results}
\end{table*}

\subsubsection{Contrastive Buffer Pairs Creation}

To effectively train a model capable of generating shades following the temporal rules (the generated shades between nearby time steps should be more similar than those with larger temporal gaps), we design a contrastive learning framework. This section introduces how we construct a training set of contrastive pairs: positive and negative pairs, based on their temporal and spatial relationships. 

Let \( D = \{ x_1, x_2, \dots, x_n \} \) represent the dataset of images, where each image \( x_i \) is associated with metadata such as its timestamp \( t_i \) and spatial location \( l_i \). For any two images \( (x_i, x_j) \), the pair is classified following equation :

\begin{equation}
    \text{Label}_{ij} = 
\begin{cases} 
1, & \text{if } l_i = l_j \text{ and } abs(|t_i - t_j|) = h, \\
0, & \text{otherwise}.
\end{cases}
\end{equation}

which describes two cases:

\textit{Positive Pair}: If \( l_i = l_j \) (same spatial location) and $abs(|t_i - t_j|) = h $  (timestamp difference of $h$ hour).

\textit{Negative Pair}: If \( l_i = l_j \) (same spatial locations) or \( abs(|t_i - t_j|) > h \) ( the difference greater than $h$ hour).

In order to ensure a balanced distribution of positive and negative pairs, for each image \( x_i \), we randomly select up to \( k_+ \) positive pairs (\( k_+ \) is a predefined maximum number of positive pairs per image\footnote{We set as 5 in our setting for learning efficiency.}). And similarly, also selects up to \( k_- \) negative pairs, ensuring sufficient variety for the contrastive learning process. The resulting dataset of pairs \( P = \{ (x_i, x_j, \text{Label}_{ij}), x_i^{edge}, x_i^{sk}, T\} \) is then used for training, the first three tuples is used for contrastive learning, and the last three is used for edge feature aggregation. For simplicity, without further clarification, the $x_i$ and $x_j$ are all satellite skeleton images in this section.


\subsubsection{Contrastive Learning for Shade Generation}

To address the challenge of maintaining temporal consistency in shade generation, we incorporate a contrastive learning framework into the training process. This framework is designed to distinguish between temporally consistent and inconsistent shade pairs by leveraging the contrastive dataset \( P = \{(x_i, x_j, \text{Label}_{ij}), x^{\text{edge}}_i, x^{\text{sk}}_i, T\} \). 

During training, for each image pair \( (x_i, x_j) \), feature embeddings \( h_i \) and \( h_j \) are obtained from a pre-trained embedding network using U-net, these embeddings are used to compute a similarity matrix \( S \in \mathbb{R}^{N \times N} \), where \( N = 2 \times B \) is the total number of embeddings in the batch (each pair contributing two embeddings), and each element \( S_{uv} \) represents the cosine similarity between embeddings as below:





\begin{equation}
    S_{uv} = \frac{h_u \cdot h_v}{\|h_u\| \cdot \|h_v\|}, \quad u, v \in \{1, \dots, N\}
\end{equation}

To optimize the temporal consistency, we apply the InfoNCE loss, which encourages embeddings of positive pairs (same location and adjacent timestamps) to be closer while pushing apart embeddings of negative pairs (differing locations or timestamps). The InfoNCE loss is expressed as:

\begin{equation}
    \mathcal{L}_{\text{contrastive}} = - \frac{1}{N} \sum_{i=1}^{N} \log \frac{\exp(S_{ii} / \tau)}{\sum_{j=1}^{N} \exp(S_{ij} / \tau)}
\end{equation}

where \( \tau \) is a temperature hyperparameter (set as 0.1 in our setting) that controls the concentration of the similarity distribution. Positive pairs (\( \text{Label}_{ij} = 1 \)) contribute to higher similarity values in \( S \), while negative pairs (\( \text{Label}_{ij} = 0 \)) are penalized to ensure their embeddings remain distinct.

The total loss function integrates the original ControlNet loss and the contrastive loss to optimize the generation of accurate and temporally consistent shade images. This is expressed as:

\begin{equation}
    \mathcal{L}_{\text{total}} = \mathcal{L}_{\text{ControlNet}} + \lambda_{\text{1}} \mathcal{L}_{\text{contrastive}}
\end{equation}

where \( \mathcal{L}_{\text{ControlNet}} \) is the primary objective for generating shade images and \( \mathcal{L}_{\text{contrastive}} \) enforces temporal consistency between embeddings of positive and negative pairs. The contrastive loss \( \mathcal{L}_{\text{contrastive}} \) is scaled by \( \lambda_{\text{1}} = 0.1 \). 

By minimizing this combined loss, the model achieves improved temporal coherence in shade generation and more nuanced outputs that align with real-world shading patterns over the time changes of a day.

\section{Experiment}
In this section, we conduct experiments to verify the effectiveness of our proposed method. We have designed three sets of experiments, the \textbf{first} is to verify the model's performance in dense building cities, such as Beijing, Phoenix downtown, and São Paulo, The \textbf{second} is to show the performance in the relatively sparse environment, such as Tempe, Brasilia, etc. Please note that these cities' selection is based on their representation in each continent, and also because they suffer from extreme heatwaves in summer, as history records. The \textbf{third}, we perform an ablation study to understand the contribution of different components in our methods.

\subsection{Metrics}
To evaluate the quality of generated shadow shades, we employ five commonly used metrics: Structural Similarity Index Measure (SSIM), Mean Squared Error (MSE), mean Intersection over Union (mIoU), Boundary Intersection over Union (B-IoU), and Learned Perceptual Image Patch Similarity (LPIPS) as in~\cite{bakurov2022structural}, ~\cite{tan2013perceptually}  ~\cite{ghazanfari2023r}. A more detailed explanation of the latter two metrics is as follows.

\begin{figure}[h!]
    \centering
    \includegraphics[width=0.99\linewidth]{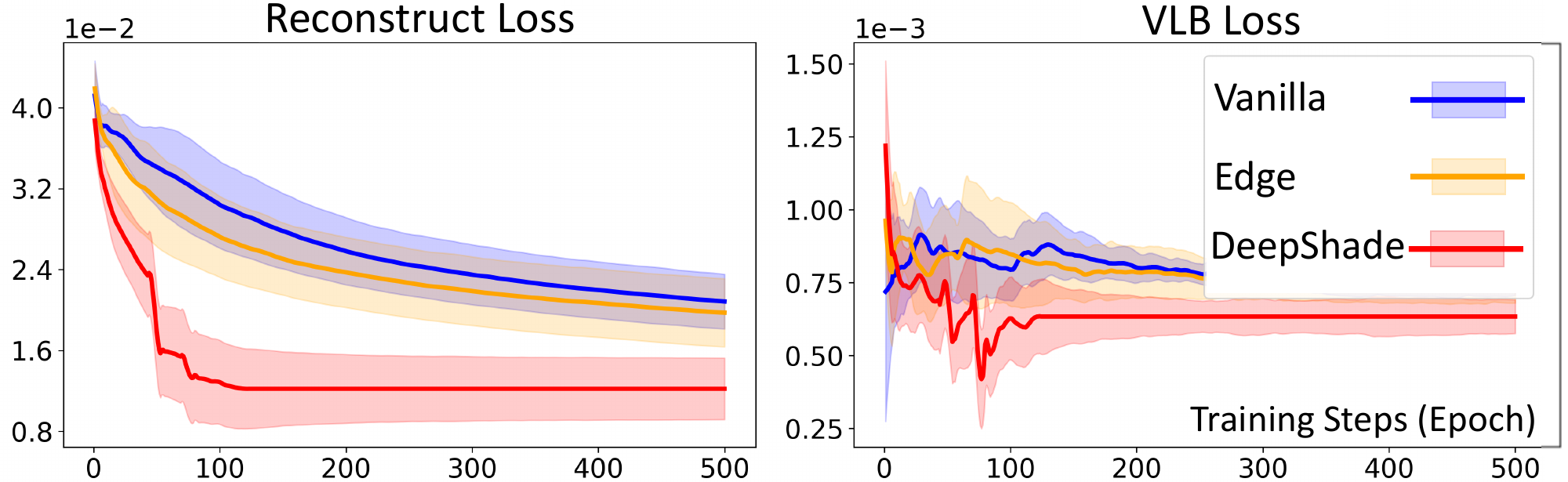}
    \caption{The training loss curves show that DeepShade shows obvious improvement regarding the convergence speed, in comparison to two baselines:  edge-conditional generation and vanilla control net. This is attributed to the integration of edge features and the contrastive framework to improve training efficiency.}
    \label{fig:loss}
\end{figure}

\paragraph{Boundary Intersection over Union (B-IoU) \(\uparrow\)}
Inspired by the work~\cite{cho2021weighted}, we proposed a new metric named: B-IoU, which measures the alignment of boundaries between the generated and ground truth shadow masks. If $M_{\mathrm{pred}}$ and $M_{\mathrm{gt}}$ are the binary shadow masks, and $K$ is a $3\times3$ structuring element. We extract their boundaries via morphological dilation and erosion:
\begin{equation}
    \partial M \;=\;\mathrm{dilate}(M, K)\;-\;\mathrm{erode}(M, K)
\end{equation}

Let's denote that $\partial M_{\mathrm{pred}}$ and $\partial M_{\mathrm{gt}}$ are the predicted and ground-truth boundaries, B-IoU is computed as the Intersection over Union of these boundary sets:

\begin{equation}
    \mathrm{B\!-\!IoU}
\;=\;
\frac{\bigl|\partial M_{\mathrm{pred}}\,\cap\,\partial M_{\mathrm{gt}}\bigr|}
{\bigl|\partial M_{\mathrm{pred}}\,\cup\,\partial M_{\mathrm{gt}}\bigr|}
\end{equation}

where $\lvert\cdot\rvert$ denotes the number of boundary pixels. The range of B-IoU is \([0,1]\), where higher values indicate better boundary alignment. Detailed implementation can be found in the evaluation code.

\paragraph{Learned Perceptual Image Patch Similarity (LPIPS) \(\downarrow\)}
LPIPS quantifies perceptual similarity by comparing feature embeddings of the generated and ground truth images extracted from a pretrained network (e.g., AlexNet). It is computed as:

\begin{equation}
    \text{LPIPS}(x, y) = \sum_{l} \frac{1}{H_l W_l} \sum_{h, w} ||\phi_l(x)_{h, w} - \phi_l(y)_{h, w}||_2^2
\end{equation}

where \(\phi_l\) denotes the feature map at layer \(l\), and \(H_l, W_l\) are its height and width. LPIPS ranges from \([0, \infty)\), where lower values indicate greater perceptual similarity.


\begin{table}[ht]
\centering
\label{tab:model_comparison}
\resizebox{0.48\textwidth}{!}{ 
\begin{tabular}{lccccc}
\toprule
\textbf{Model}& \textbf{SSIM\(\uparrow\)} & \textbf{mIoU\(\uparrow\)}& \textbf{B-IoU\(\uparrow\)}& \textbf{MSE\(\downarrow\)} & \textbf{LPIPS\(\downarrow\)} \\ 
\midrule
Backbone Model (direct) &
0.4252$_{\pm\text{0.01}}$ & 0.0358$_{\pm\text{0.00}}$ &
0.0213$_{\pm\text{0.00}}$  & 41.2666$_{\pm\text{1.65}}$ &  0.7967$_{\pm\text{0.00}}$              \\ 
Vanilla Control Net &  0.9690$_{\pm\text{0.04}}$ & 0.2736$_{\pm\text{0.13}}$ &  0.0812$_{\pm\text{0.05}}$ & 18.3388$_{\pm\text{3.37}}$&          0.3304$_{\pm\text{0.03}}$     \\ 
Edge Condition & 0.9684$_{\pm\text{0.01}}$ & 0.2898$_{\pm\text{0.04}}$ & 0.1040$_{\pm\text{0.01}}$ &
18.6686$_{\pm\text{0.70}}$ &  0.3358$_{\pm\text{0.01}}$               \\ 
\textbf{Ours (DeepShade)}                  &      \textbf{0.9692}$_{\pm\text{0.04}}$   &  \textbf{0.2903}$_{\pm\text{0.20}}$  & \textbf{0.1240}$_{\pm\text{0.07}}$  & \textbf{18.1721}$_{\pm\text{4.09}}$ &  \textbf{0.3024}$_{\pm\text{0.29}}$  \\ 
\bottomrule
\end{tabular}
} 
\caption{Comparison of different models across various metrics: SSIM, MSE, mIoU, LPIPS, and B-IoU. In this experiment, each of the trained models is fed with a bird’s-eye view satellite skeleton image and a text prompt describing the time and solar angle, and our method consistently performs better than these baseline methods.}
\end{table}

\begin{figure*}[t!]
    \centering
    \includegraphics[width=0.98\linewidth]{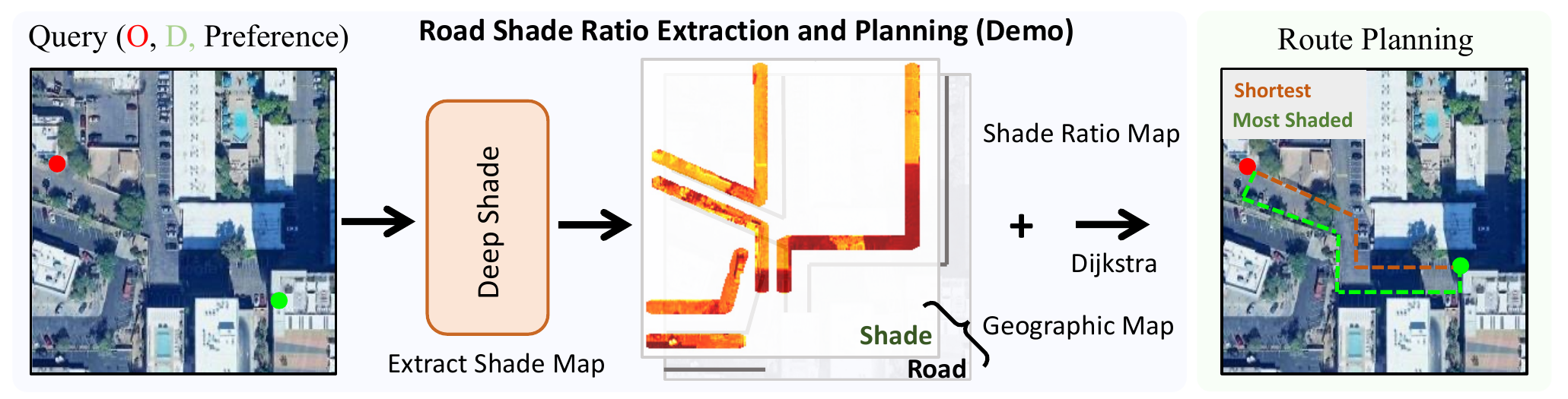}
    \caption{The demo of using our trained DeepShade in Tempe city for shade ratio extraction based on the generated shade maps, and we can effectively leverage this information to conduct planning that balances the distance and shade exposure at the Noon of a day (the preference score of shade is set as 50\%, so distance and shade take half of the weight during planning).}
    \label{fig:demo}
\end{figure*}
\subsection{Experiment Analysis}
\paragraph{A. The Efficient Convergence Speed.} As shown in Figure.~\ref{fig:loss}, the plotted curves are the reflection of the mean and standard deviation of the 5-rounds result for each method across two commonly adopted losses (reconstruction and VLB loss~\cite{deja2022analyzing}). 
Our model DeepShade demonstrates a much more efficient convergence performance in comparison to baseline methods: edge-condition diffusion and vanilla controller. The training of diffusion models is essentially difficult; however, our proposed method provides a more rational way in the shade prediction setting.

\paragraph{B. Accurate Generation Across 12 Cities in the World.}
In this part, we focus on the Table.~\ref{tab:results}. The upper part is the dense areas in the world selected dataset for testing, and the lower part is the sparse areas. We can observe that \ our model outperforms most of the baseline methods in the various metrics (with a total of 50 epochs of training), regardless of the dense or relatively sparse scenario land covers. It demonstrates the vivid simulation of the conditioned area using the ControlNet-based method by training on one dataset, with good transferability, it better indicates that this work has great potential for real-world applications.

\paragraph{C. Ablation Study.}  We also included an ablation study in the paper, as in Table 2, it is the training conducted in the Tempe dataset, and the test is the other 30\% from the original dataset given by the default split. This result reveals the importance of each component in our method. We can see that the edge condition (our method without contrastive learning) suffers the most performance drop in comparison to the DeepShape full model structure, and if we also remove the edge conditions, the performance further decreases, the backbone model is the stable diffusion model (all of the above models are trained 5 times with mean and std reported).


\section{Demonstration}
Besides the quantitative analysis in experiments, it is important to demonstrate the real-world impact of the work. Thus, we design a proof-of-concept demo as shown in Figure.~\ref{fig:demo} using a subarea of Arizona State University. In this demo, we tackle the problem of integrating the shade ratio as a factor when making the routing suggestions. The input is a skeleton image containing the building outline, extracted from a satellite image of the interested area. Then, based on the time for planning, we describe as the text prompt $T$ together with the image that is processed to $x^{cond}$ as in Eq.~\ref{eq:input}, the shape map will be generated as output, it will be used for shade ratio calculation by overlaying the shade map with the road using longitude and latitude ranges. Given the shade ratio, the planning is made by jointly considering the user's preference (weight) on shade and distance by a variant of the Dijkstra algorithm. The green shows a more shaded plan, while the red means the shortest path. This demo reveals the potential of a real-world application; given that shade-involved planning is crucial for areas that suffer from extreme heat waves, this demonstration shows a way that could possibly help decrease heat stroke cases and improve the health of outdoor people.

\section{Conclusion}
In this work, we focus on the challenge of simulating realistic urban shade patterns by introducing both a novel dataset and a generative framework. We first developed a globally diverse dataset of building layouts with aligned satellite imagery and timestamped shade snapshots, paired with text‐conditioned prompts encoding solar zenith angle and time of day. We then propose \ours, a text‐conditioned, edge‐enhanced diffusion model built on ControlNet that fuses building skeletons and Canny edges and incorporates a contrastive learning module to enforce temporal consistency. We have conducted extensive experiments across multiple cities, showing that \ours could function reasonably well in in‐domain and out‐of‐domain tests. An ablation study further validates the importance of edge conditioning and contrastive loss, and lastly, we show a proof‐of‐concept application in shaded route planning, which demonstrates the work's practical utility for heat‐aware urban navigation. The dataset~\footnote{\url{https://huggingface.co/datasets/DARL-ASU/DeepShade}} and codebase~\footnote{\url{https://github.com/LongchaoDa/DeepShade_repo.git}} are released for the promotion of research in this direction.

\section{Limitation and Future Work}
The paper proposes a very first dataset that aligns shades to the satellite view of a location; however, there might be missing building issues in ground truth shades since the OSM data is not always updated as the time that the satellite image is taken. Future development to tackle this problem is important; meanwhile, the real-time shade planning would require the pre-generated shade ratios, and they should be regularly indexed to the real-world road maps to enable large-scale support for the planning systems. A more accurate and robust framework is worth exploring.

\section*{Acknowledgments}
The work was partially supported by NSF awards \#2421839, NAIRR \#240120, and used AWS through the CloudBank project, which is supported by NSF grant \#1925001. Xiangrui was supported by the Fulton Fellowship at ASU. The views and conclusions in this paper are those of the authors and should not be interpreted as representing any funding agencies. Thank OpenAI for providing API credits under the Researcher Access program and Amazon Research Awards.

\section*{Contribution Statement }
Longchao Da and Xiangrui Liu both contributed equally to the paper.




\bibliographystyle{named}
\bibliography{ijcai25}
\clearpage
\appendix

\end{document}